\documentclass{article}
\usepackage[utf8]{inputenc}
\usepackage{authblk}
\usepackage{natbib}
\usepackage[a4paper, total={6in,10in}]{geometry}

\providecommand{\keywords}[1]
{
  \small	
  \textbf{\textit{Keywords:}} #1
}
\title{Metaheuristic for Hub-Spoke Facility Location Problem: Application to Indian E-commerce Industry}

\author[1]{Aakash Sachdeva}

\author[2]{Bhupinder Singh}

\author[3]{Rahul Prasad}

\author[4]{Nakshatra Goel}

\author[5]{Ronit Mondal}

\author[6]{Jatin Munjal}

\author[7]{Abhishek Bhatnagar}

\author[8]{Manjeet Dahiya}
\affil[1,2,3,4,5,6,7,8]{Data Sciences department, Ecom Express Limited, 10th floor, Ambience Tower II, Ambience Island,Gurugram,122022,Haryana, India}

\begin{document}

\maketitle
\begin{center}
\textbf{\large Abstract}  
\end{center}

\large
Indian e-commerce industry has evolved over the last decade and is expected to grow at 19\% \citep{IBEF2022} per year over the next 3 years. COVID-19 pandemic induced lockdown has been a key driver in this surge; this trend is set to continue over the next few years \citep{pages2019designing}. The focus has now shifted on turn around time (TAT) due to emergence of many third-party logistics providers and higher customer expectations. TAT refers to the time from when the customer manifests an order to the time it gets delivered. The key consideration for delivery providers is to balance their overall operating costs along with meeting the promised TAT to their customers \citep{Anvari2017}.

E-commerce delivery partners operate through a network of facilities \citep{verma2022data}. These facilities include (i) pick-up centers: these are mapped to hubs and send the outbound shipments to be bagged and moved through the hubs to the destination city; (ii) fulfilment centers (FC): these centers maintain stocks of highly sought out products for faster delivery; (iii) pre-processing centers (PPC): these facilities perform bagging of shipments received from pick-up and fulfilment centers; (iv) hubs: these serve as consolidation points for shipment bags (often co-located with PPC) usually located in different cities, typically low in number but large in capacity; and (v) distribution centers (DC): these are smaller facilities, large in number and mapped to a unique hub from where it receives the inbound bags for the destination city.

E-commerce logistics operations are broadly classified into three stages, namely \emph{first mile}, \emph{middle mile} and \emph{last mile} \citep{laseinde2017providing}. First and last mile operations are typically intra-city operations, while the middle mile operations are inter-city operations. A first mile operation starts when a field executive picks up the customer shipment from a pick-up/ fulfilment center and delivers it to pre-processing center (PPC). PPC bags this shipment and then sends the bag to the hub in the origin city (co-located with the PPC). The bag then moves to the destination city hub. Hub sorting operations help to segregate the packages based on their PIN code and localities. Each PIN code/ locality is mapped to a unique DC, which receives the bags from the hub. The entire operations, from the origin city hub to the destination city DC, are known as middle mile operations. The last mile operations require the field executives to pickup the shipments from the DC and deliver them to the consignee. A huge component of the transportation costs incur in the middle mile operations and subsequently offers a good opportunity for savings through strategic location of hubs and their appropriate mapping with the distribution centers.

E-commerce industry works on the principle of economies of scale where shipments are consolidated at hubs before moving to other hubs, and then to the DCs. Hub to hub movement of goods are through line hauls, a widely used concept in logistics that aids consolidation of goods for various destinations to improve efficiency and lower the transportation costs \citep{choi2021comparison}. Hub to DC movements are through milk runs \citep{uygun2022ant}, where a set of DCs are a part of a vehicle route based on available capacity. A milk run is a route starting from a hub, visiting designated DCs in a predetermined sequence and schedule, and ending at the same hub.

Operations research has been at the forefront of logistics to aid in a variety of decision making \citep{Farahani2019}. The key decisions include the locations of facilities such as hubs and distribution centers; routing and scheduling of line hauls and milk runs; and manpower planning \citep{Anvari2017}. In this work, we solve the facility location problem for hubs and their corresponding mappings with the DCs. We transform the problem into discrete domain where all DCs are potential locations for the hubs. The decision variables ascertain whether a given DC can be an optimal location for a hub in a network of facilities and corresponding sorter capacity recommendations. The model considers shipments flow direction; inbound and outbound at a DC to optimally locate hubs and map a DC to it. We model the facility location problem for a five-year forward looking perspective due to its strategic nature \citep{arabani2012facility}.

The focus of delivery partners has now shifted towards delivering goods faster rather than only through the minimum cost transport mode, as the case earlier. Generally, metro cities have faster TAT compared to non-metro counterparts due to availability of good transport infrastructure and high demand. This encourages logistics providers to plan line hauls to prioritize high demand locations and good TAT compliance. It should be noted that the operations should be cost efficient while incorporating the customer TAT requirements. Hence, it is imperative to find best locations to serve these objectives \citep{celik2020comparative}. 

It is not manually possible to take location and allocation decisions given the large network size and associated constraints. This needs systematic mathematical modelling, however, it is not possible to solve such a problem using exact optimization methods. There is a need to employ a metaheuristic based solution to solve a large scale problem efficiently as done in the current study. The underlying modelling concept is based on the business requirement to deliver goods at a DC from the destination side hub within 24 hours of arrival at the hub. The aim is to minimize the overall network cost that includes hub setup costs, line haul costs, milk run costs and TAT breaches/ penalty costs. Hub setup costs include hub floor area cost, handling cost and sorter cost. Line haul cost is the cost associated with movement of shipments between hubs. Milk run cost has two components: one component each on the arrival and the departure side. Arrival side cost consists of movement between origin side pick-up/ fulfilment center to the hub, whereas, departure side milk run costs are the movement cost from destination side hub to DC. Penalty cost is the potential TAT breach cost throughout the middle mile movement of shipments. This component is usually based on direct origin destination distance and travel time calculated based on feasibility of a given amount of driving distance per day. The objective function aims to minimize total costs which is the sum of these costs with proper scale factors computed using historical data. Typically, the hubs are located to cater to the peak demand during the festive season in India (around October-November every year).

We use Genetic Algorithm (GA) for this purpose. GA is a biologically inspired metaheurestic that uses processes such as selection, crossover and mutation to return a solution for a given population size and number of generations \citep{atta2019multi, Dwivedi2020}. We leverage business constraints to reduce the computation time of GA, a key contribution of our study. The solution methodology takes the input as origin pickup/ fulfilment center and destination DC shipment volumes that help calculate the flow volumes; geographical coordinates of all DCs to compute the distance matrix between facilities; list of already existing hubs that cannot close, hence, assumed to be fixed (helps reduce the solution time). We consider road distances in our analysis. The output are a set of hubs that will operate, their sorter configurations, hub floor area and corresponding hub-DC mappings. The parameters include different types of costs mentioned above. Further, to keep the solution search space within realistic bounds, we provide a tentative range of hubs to be kept operational, minimum number of DCs that should be mapped to a hub, minimum volume to be mapped to a hub and minimum inter hub distance (not applicable to already fixed hubs). 

We apply this to a real-world problem of a large e-commerce logistics provider in India. The provider delivers 10 lakh shipments daily on an average throughout the year, which rises to 2.5-3 times during the festive season. We design the facilities network for this peak load. The results show considerable cost and TAT benefits assuming these are feasible on ground. We perform multiple simulations and report two key results as follows: 
\emph{simulation-01} gives optimal location of 76 hubs, however, this design has ground implementation challenges due to long milk run designs. To overcome this, we multiply both arrival and departure side milk run costs with a scale factor to normalize milk run costs with respect to hub setup costs. This gives rise to \emph{simulation-02} with 94 hubs and no milk run breaches. Further, to rationalize the model output with the current on-ground situation and minimize the operational disruptions and labour resistance, we recommend to keep some current hubs active that are within a stipulated distance (approx. 50 kms) from the simulation recommended hubs. This recommendation includes closing 16 existing hubs and opening 17 hubs at new locations for FY-2023 with a total of 82 hubs in the network, an increase of 1 hub from the current scenario. Other hubs may be commissioned in the subsequent years to ensure ease of implementation.  The results with 82 hubs is able to reduce daily TAT breaches by 9.73\% as compared to the current scenario.The new locations will be able to cater to the predicted volume surge over the next five-years. Given the scale of operations and the problem complexity, this exercise is carried out twice a year:(i) at least six-months before the festive season months to ensure timely implementation of the rationalized plan and (ii) immediately after the festive season month ends to understand the gaps between the implemented plan based on forecast and the actual festive season volumes. Overall, this is a beneficial activity for the organization, which is manually intractable.

\keywords\large{Facility location problem, hub-spoke model, genetic algorithm, turn around time}

\section*{Acknowledgements}
\large
The authors would like to thank Ecom Express Limited and Mr. T.A. Krishnan (CEO, Ecom Express) in particular for his continuous support. The authors would like to thank Mr. Sonam Paliwal (Head, Network Operations), Mr. Krishnananda (Head, Planning and Engineering), Mr. Vishwachetan Nadamani (Chief Operating Officer), Mr. Amit Choudhary (Chief Product Technology Officer), Mr. Prashant Gazipur (Country Head of Operations) and Mr. Santosh Rawat (Ex-Senior Manager) at Ecom Express Limited for insighful discussions. 

\bibliographystyle{apalike}
\bibliography{bibliography.bib}
\end{document}